\documentclass[10pt, twocolumn]{article}

\usepackage[utf8]{inputenc}
\usepackage{amsmath, amssymb, amsthm}
\usepackage{graphicx}
\usepackage{booktabs}
\usepackage{float}
\usepackage{hyperref}
\usepackage{geometry}
\usepackage{caption}
\usepackage{subcaption}

\usepackage{tikz}
\usetikzlibrary{
    positioning,
    arrows.meta,
    shapes.geometric,
    calc,
    fit,
    backgrounds
}
\usetikzlibrary{decorations.pathreplacing}

\usepackage{abstract}

\title{\textbf{BiJEPA: Bi-directional Joint Embedding Predictive Architecture for Symmetric Representation Learning}}
\author{Yongchao Huang\footnote{Author email: yongchao.huang@abdn.ac.uk}}
\date{08/02/2026}

\begin{document}

\maketitle

\begin{abstract}
Self-Supervised Learning (SSL) has shifted from pixel-level reconstruction to latent space prediction, spearheaded by the Joint Embedding Predictive Architecture (JEPA). While effective, standard JEPA models typically rely on a uni-directional prediction mechanism (e.g. Context $\to$ Target), potentially neglecting the informative signal inherent in the inverse relationship, degrading its performance. In this work, we propose \textbf{BiJEPA}, a \textit{Bi-Directional Joint Embedding Predictive Architecture} that enforces cycle-consistent predictability between data segments. We address the inherent instability of symmetric prediction (representation explosion) by introducing a critical norm regularization mechanism on the representation vectors. We evaluate BiJEPA on three distinct modalities: synthetic periodic signals, chaotic Lorenz attractor trajectories, and high-dimensional image data (MNIST). Our results demonstrate that BiJEPA achieves stable convergence without collapse, captures the semantic structure of chaotic systems, and learns robust temporal and spatial representations capable of generation and generalisation, offering a more holistic approach to representation learning.
\end{abstract}

% ----------------------------------------------------------------------
\section{Introduction}
\label{sec:intro}

The fundamental goal of Self-Supervised Learning (SSL) is to learn useful representations of data without human annotation \cite{gui2023survey}. Generative methods (e.g., Autoencoders \cite{hinton2006reducing}, MAE \cite{he2022masked}) achieve this by reconstructing raw input pixels, often wasting capacity on high-frequency noise. In contrast, Joint Embedding Architectures (JEAs), such as VICReg \cite{bardes2022vicreg} and SimCLR \cite{chen2020simple}, focus on invariance, pulling views of the same object together.

The \textit{Joint Embedding Predictive Architecture} (JEPA) occupies a middle ground: it learns to predict the \textit{representation} of missing information from available context \cite{lecun2022path}. Unlike generative models (e.g., MAE, Autoencoders) which reconstruct high-frequency pixel details, JEPA predicts in abstract space. Conversely, unlike invariance models (e.g., SimCLR, VICReg) which aim to make $f(x)$ identical to $f(y)$ by collapsing views together, JEPA enforces a \textit{predictive} relationship. It acknowledges that $x$ and $y$ are distinct states (e.g., past \textit{vs.} future) and learns the transformation that maps one to the other, rather than forcing them to identical points in the embedding space, thereby encouraging representation diversity.

However, standard JEPA implementations (e.g., I-JEPA) are fundamentally uni-directional. They designate one view as the ``Context'' ($x$) and another as the ``Target'' ($y$), training a predictor $P$ such that $P(f_\theta(x)) \approx f_{\bar{\theta}}(y)$. However, in many physical and semantic systems, the relationship is bi-directional. In temporal data, $t \to t+1$ follows causal physics, while $t+1 \to t$ follows inverse dynamics. In spatial data, the left side of an object implies the right, and vice versa. Ignoring the backward pass wastes half the available supervisory signal.

\paragraph{Novelty and contributions.}
We introduce \textbf{BiJEPA}, a symmetric architecture that trains two \textit{distinct} \footnote{If needed, the context and target can share a predictor using e.g. invertible neural net. See later discussions in Section \ref{subsec:shared_predictor}.} predictors simultaneously: Forward ($x \to y$) and Backward ($y \to x$). Our contributions are:
(i) \textit{Symmetric Architecture:} a dual-predictor framework that learns reversible semantic mappings.
(ii) \textit{Stability Analysis:} we identify ``Representation Explosion'' as a primary failure mode of bi-directional SSL and demonstrate that effective norm regulation (via soft or hard constraints) is a necessary condition for convergence.
(iii) \textit{Generative Validation:} we propose a ``Generative Decoder'' probe to visually verify that the embeddings retain sufficient geometric information to hallucinate missing data.

% ----------------------------------------------------------------------
\section{Related Work}
\label{sec:related}

Since the introduction of the JEPA paradigm \cite{lecun2022path}, several variants have emerged, tailoring the architecture to specific domains and control problems.
\textit{H-JEPA (Hierarchical JEPA)} \cite{lecun2022path} introduces multi-scale processing, learning representations at varying levels of granularity to capture global structure, providing a normative blueprint for predictive world modeling and hierarchical control without observation reconstruction.
\textit{I-JEPA (Image-based JEPA)} \cite{assran2023ijepa} applies the framework to vision transformers, predicting the embeddings of masked target blocks from context blocks. It relies on an asymmetric Exponential Moving Average (EMA \cite{mnih2015human,grill2020byol}) target encoder to prevent collapse.
\textit{V-JEPA (Video JEPA)} \cite{bardes2024vjepa,bardes2025VJEPA2} extends this to the temporal domain, focusing on strictly causal $t \to t+k$ prediction to learn motion dynamics. Recent work has demonstrated that such large-scale pretraining enables zero-shot robotic manipulation via latent-space model-predictive control \cite{bardes2025VJEPA2}.

\textbf{World models and planning.} Building on this vision, recent research has formalized JEPA as a foundation for planning. \textit{JEPA World Models (JEPA-WMs)} \cite{terver2025jepaworldmodels} combine pretrained visual encoders with action-conditioned predictors, enabling planning by optimizing action sequences directly in the latent space. \textit{Value-Guided JEPA} \cite{Destrade2026ValueGuidedJEPA} shapes the representation space such that distances approximate goal-conditioned value functions, thereby improving the optimization landscape. Most recently, \textit{VJEPA} \cite{huang2026vjepavariationaljointembedding} formulates the architecture as a \textit{probabilistic} world model, using variational inference to capture uncertainty in latent predictions, contrasting with the deterministic objectives of prior works.

\textbf{Domain adaptations.} Beyond these foundational instantiations, the framework has inspired a variety of domain-specific adaptations. \textit{MC-JEPA} \cite{bardes2023mcjepa} explicitly decouples motion and content. \textit{S-JEPA} \cite{abdelfattah2024sjepa} applies the principle to skeletal action recognition, while \textit{Point-JEPA} \cite{saito2025pointjepa} adapts it for point clouds. Other variants explore JEPA in graph domains (\textit{Graph-JEPA} \cite{Skenderi2025graphJEPA}), audio spectrograms (\textit{Audio-JEPA} \cite{tuncay2025audiojepa}), and joint vision-language models (\textit{VL-JEPA} \cite{chen2025vlJEPA}). Additional work, such as \textit{C-JEPA} \cite{mo2024cJEPA} and \textit{DSeq-JEPA} \cite{he2025dseqJEPA}, proposes contrastive or sequential extensions to address specific limitations like representation collapse.

\textbf{Motivation.} While these models excel in their domains, they predominantly rely on uni-directional prediction strategies. BiJEPA distinguishes itself by explicitly modeling the inverse relationship. This design is directly inspired by physical phenomena such as reversible wave propagation, where prior work by Huang et al. (2023) demonstrated the efficacy of enforcing consistency via forward and backward physical losses in neural-physical models \cite{huang2023bayesian}. Just as their framework utilizes the spatial reversibility of wave dynamics to constrain the loss landscape, BiJEPA adapts this principle to the latent embedding space, moving beyond simple uni-directional prediction to enforce rigorous symmetric predictability.

% ----------------------------------------------------------------------
\section{Methodology}
\label{sec:method}

\subsection{Classic JEPA Architecture}

The Joint Embedding Predictive Architecture (JEPA) posits that useful representations can be learned by predicting the embeddings of missing or future data, rather than reconstructing high-entropy observations (e.g. pixels) \cite{lecun2022path}. The central design principle is to encourage representations that capture \textit{predictable, task-relevant structure} while discarding nuisance variability that is difficult or unnecessary to predict \cite{huang2026vjepavariationaljointembedding}.

\begin{figure}[t]
    \centering
    \resizebox{\linewidth}{!}{\begin{tikzpicture}[
    node distance=1.0cm and 1.5cm,
    >=Stealth,
    font=\small\sffamily,
    % --- Styles ---
    block/.style={draw, rectangle, rounded corners, minimum width=1.6cm, minimum height=0.9cm, align=center, fill=white, line width=0.8pt},
    encoder/.style={block, fill=blue!5, draw=blue!60!black},
    predictor/.style={block, fill=orange!5, draw=orange!60!black},
    target_enc/.style={block, fill=gray!5, draw=gray!60!black, dashed},
    state/.style={circle, draw, minimum size=0.9cm, inner sep=0pt, fill=white, line width=0.8pt, font=\footnotesize},
    loss_node/.style={diamond, draw=red, fill=red!5, inner sep=1pt, aspect=1.5, font=\scriptsize},
    label_text/.style={font=\bfseries, align=left},
    flow_line/.style={->, line width=0.8pt},
    stop_grad/.style={midway, font=\tiny, text=gray, fill=white, inner sep=1.5pt}
]

% =================================================================================
% --- A. CLASSIC JEPA ---
% =================================================================================
\node[label_text] (label_a) at (0, 0) {(A) Classic JEPA};

% -- 1. Online Branch --
\node[below=1.0cm of label_a.west, anchor=west] (x_a) {Input $x$};
\node[encoder, right=0.8cm of x_a] (enc_online_a) {Online\\$f_\theta$};
\node[state, right=0.8cm of enc_online_a] (sx_a) {$s_x$};
\node[predictor, right=0.8cm of sx_a] (pred_a) {Predictor\\$P_\phi$};
\node[state, right=0.8cm of pred_a] (sy_hat_a) {$\hat{s}_y$};

% Latent Z
\node[above=0.5cm of pred_a, font=\footnotesize] (z_a) {$z$};
\draw[->] (z_a) -- (pred_a);

% -- 2. Target Branch --
\node[below=1.5cm of x_a] (y_a) {Target $y$};
\node[target_enc, right=0.8cm of y_a] (enc_target_a) {Target\\$f_{\bar{\theta}}$};
% State positioned relative to prediction for alignment
\node[state] (sy_a) at (sy_hat_a |- enc_target_a) {$s_y$};

% Connections A
\draw[flow_line] (x_a) -- (enc_online_a);
\draw[flow_line] (enc_online_a) -- (sx_a);
\draw[flow_line] (sx_a) -- (pred_a);
\draw[flow_line] (pred_a) -- (sy_hat_a);
\draw[flow_line] (y_a) -- (enc_target_a);
\draw[flow_line] (enc_target_a) -- node[stop_grad] {SG} (sy_a);
\draw[<->, dashed, red, thick] (sy_hat_a) -- node[midway, right, font=\scriptsize, text=red] {Loss} (sy_a);

% =================================================================================
% --- SEPARATOR ---
% =================================================================================
\draw[thick, gray!30] ($(y_a.south west)+(-0.5,-0.8)$) -- ($(sy_hat_a.east |- y_a.south east)+(2.5,-0.8)$);

% =================================================================================
% --- B. BiJEPA (Ours) ---
% =================================================================================
\node[label_text, below=1.6cm of y_a.west, anchor=west] (label_b) {(B) BiJEPA (Ours)};

% ------------------------------------------------------------------
% SUB-DIAGRAM 1: Forward Process (x -> y)
% ------------------------------------------------------------------
\node[below=0.8cm of label_b.west, anchor=west] (bx_fwd) {Input $x$};
\node[encoder, right=0.8cm of bx_fwd] (b_enc_x) {Online\\$f_\theta$};
\node[state, right=0.8cm of b_enc_x] (b_sx) {$s_x$};
\node[predictor, right=0.8cm of b_sx, minimum width=1.6cm] (b_pred_fwd) {$P_{fwd}$};
\node[state, right=0.8cm of b_pred_fwd] (b_sy_hat) {$\hat{s}_y$};

% Latent z1
\node[above=0.5cm of b_pred_fwd, font=\footnotesize] (bz1) {$z_1$}; \draw[->] (bz1) -- (b_pred_fwd);

% Target Branch (Local) 
\node[below=1.5cm of bx_fwd] (by_tgt) {Target $y$};
\node[target_enc, right=0.8cm of by_tgt] (b_enc_y_tgt) {Target\\$f_{\bar{\theta}}$};
% Target State (Right side, aligned with prediction)
\node[state] (b_sy_ref) at (b_sy_hat |- b_enc_y_tgt) {$s_y^{tgt}$};

% Connections Fwd
\draw[flow_line] (bx_fwd) -- (b_enc_x);
\draw[flow_line] (b_enc_x) -- (b_sx);
\draw[flow_line] (b_sx) -- (b_pred_fwd);
\draw[flow_line] (b_pred_fwd) -- (b_sy_hat);
% Long Target Arrow
\draw[flow_line] (by_tgt) -- (b_enc_y_tgt);
\draw[flow_line] (b_enc_y_tgt) -- node[stop_grad] {SG} (b_sy_ref);

% Loss Fwd (Vertical Arrow)
\draw[<->, dashed, red, thick] (b_sy_hat) -- node[midway, right, font=\scriptsize, text=red, fill=white, inner sep=1pt] {$L_{fwd}$} (b_sy_ref);

% ------------------------------------------------------------------
% SUB-DIAGRAM 2: Backward Process (y -> x)
% ------------------------------------------------------------------
% CHANGED: Reduced from 3.5cm to 3.0cm to tighten the layout
\node[below=3.0cm of bx_fwd] (by_bwd) {Input $y$};
\node[encoder, right=0.8cm of by_bwd] (b_enc_y) {Online\\$f_\theta$};
\node[state, right=0.8cm of b_enc_y] (b_sy) {$s_y$};
\node[predictor, right=0.8cm of b_sy, minimum width=1.6cm] (b_pred_bwd) {$P_{bwd}$};
\node[state, right=0.8cm of b_pred_bwd] (b_sx_hat) {$\hat{s}_x$};

% Latent z2
\node[below=0.5cm of b_pred_bwd, font=\footnotesize] (bz2) {$z_2$}; \draw[->] (bz2) -- (b_pred_bwd);

% Target Branch (Local)
\node[below=1.5cm of by_bwd] (bx_tgt) {Target $x$};
\node[target_enc, right=0.8cm of bx_tgt] (b_enc_x_tgt) {Target\\$f_{\bar{\theta}}$};
% Target State (Right side)
\node[state] (b_sx_ref) at (b_sx_hat |- b_enc_x_tgt) {$s_x^{tgt}$};

% Connections Bwd
\draw[flow_line] (by_bwd) -- (b_enc_y);
\draw[flow_line] (b_enc_y) -- (b_sy);
\draw[flow_line] (b_sy) -- (b_pred_bwd);
\draw[flow_line] (b_pred_bwd) -- (b_sx_hat);
% Long Target Arrow
\draw[flow_line] (bx_tgt) -- (b_enc_x_tgt);
\draw[flow_line] (b_enc_x_tgt) -- node[stop_grad] {SG} (b_sx_ref);

% Loss Bwd (Vertical Arrow)
\draw[<->, dashed, red, thick] (b_sx_hat) -- node[midway, right, font=\scriptsize, text=red, fill=white, inner sep=1pt] {$L_{bwd}$} (b_sx_ref);

% ------------------------------------------------------------------
% FINAL JOINT LOSS (Right Side)
% ------------------------------------------------------------------
% Create a node encompassing both loss locations to draw the brace
\node (top_loss) at ($(b_sy_hat)!0.5!(b_sy_ref)$) {};
\node (btm_loss) at ($(b_sx_hat)!0.5!(b_sx_ref)$) {};

% Big Brace
% \draw[decorate, decoration={brace, amplitude=10pt, mirror}, gray, line width=1pt] 
%     ($(top_loss.east)+(1.4, 0.5)$) -- ($(btm_loss.east)+(1.4, -0.5)$)
%     node[midway, right=15pt, align=center, font=\bfseries\small, text=black] (total_loss) {Total Loss\\$L_{total}$};
\draw[decorate, decoration={brace, amplitude=10pt}, gray, line width=1pt] 
    ($(top_loss.east)+(1.4, 0.5)$) -- ($(btm_loss.east)+(1.4, -0.5)$)
    node[midway, right=15pt, align=center, font=\bfseries\small, text=black] (total_loss) {Total Loss\\$L_{total}$};

% Arrows from individual losses to the brace
\draw[->, dotted, thick, gray] (top_loss) -- ($(top_loss.east)+(1.4,0)$);
\draw[->, dotted, thick, gray] (btm_loss) -- ($(btm_loss.east)+(1.4,0)$);

\end{tikzpicture}}
    \caption{JEPA \textit{vs.} BiJEPA. (A) Standard JEPA learns a uni-directional mapping ($x \to y$). (B) BiJEPA adds a backward predictor $P_{bwd}$ and enforces consistency in both directions (symmetric predictability), learning to map $x \to y$ and $y \to x$ simultaneously using Online ($f_\theta$) and Target ($f_{\bar{\theta}}$) encoders to prevent collapse. SG denotes the Stop-Gradient operation. NB: both the forward and backward loops share the same Online encoder $f_{\theta}$ and Target encoder $f_{\bar{\theta}}$.}
    \label{fig:arch_comparison}
\end{figure}
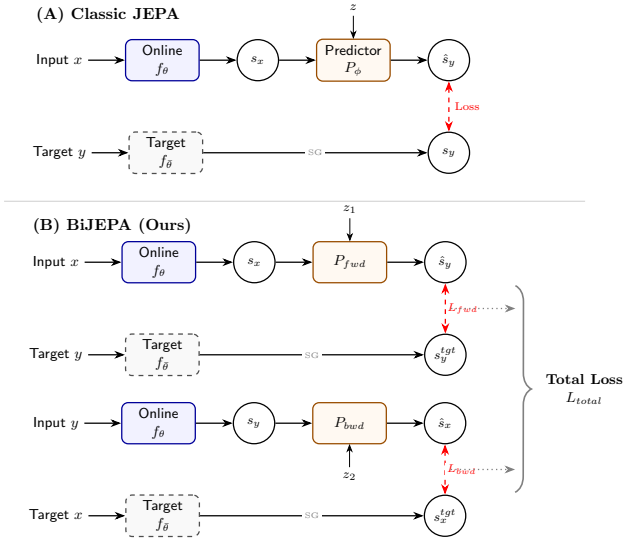

The classic JEPA is shown in Fig.\ref{fig:arch_comparison}. Given an input data stream, JEPA partitions the data into a \textbf{Context} $x$ and a \textbf{Target} $y$. In spatial domains (e.g., I-JEPA), this corresponds to masking patches of an image; in temporal domains (e.g., V-JEPA), $x$ represents past frames and $y$ represents future frames.
The context is mapped to a latent representation $s_x$ via a trainable \textbf{Online Encoder} $f_\theta$:
\begin{equation}
    s_x = f_\theta(x).
\end{equation}
Simultaneously, the target is processed by a separate \textbf{Target Encoder} $f_{\bar{\theta}}$ to produce the semantic training target $s_y$:
\begin{equation}
    s_y = f_{\bar{\theta}}(y).
\end{equation}
To predict this target representation, a \textbf{Predictor} network $P_\phi$ maps the context embedding $s_x$, conditioned on a variable $z$ (which may encode stochasticity or structural side information like mask positions), to a prediction $\hat{s}_y$:
\begin{equation}
    \hat{s}_y = P_\phi(s_x, z).
\end{equation}
The system is trained by minimizing the distance between the predicted and actual target representations:
\begin{equation}
    L = || \hat{s}_y - s_y ||^2_2 = || P_\phi(f_\theta(x), z) - f_{\bar{\theta}}(y) ||^2_2.
\end{equation}

\paragraph{Optimization and Collapse Prevention}
To ensure the learning of non-trivial representations, JEPA relies on a specific optimization asymmetry. If gradients flowed through both encoders, the model could trivially minimize the loss by outputting constant vectors (e.g., all zeros) - a failure mode known as \textit{representation collapse}.
To prevent this, gradients derived from the loss are stopped before reaching the Target Encoder. Consequently, the weights of the Target Encoder $\bar{\theta}$ are not updated via backpropagation but are instead evolved via an \textit{Exponential Moving Average} (EMA) of the Online Encoder's weights $\theta$ \cite{mnih2015human,mnih2016asynchronousmethodsdeepreinforcement,lillicrap2019continuouscontroldeepreinforcement,grill2020byol}:
\begin{equation} \label{eq:ema_update}
    \bar{\theta} \leftarrow \tau \bar{\theta} + (1-\tau)\theta,
\end{equation}
where $\tau \in [0, 1]$ is a momentum coefficient. This mechanism ensures that $s_y$ provides a stable, slowly drifting regression target, forcing the Online Encoder and Predictor to actively learn semantic features to match the target, rather than the target collapsing to match the prediction.

\subsection{Bi-Directional JEPA (BiJEPA)}

Unlike standard JEPA, which enforces a strictly uni-directional predictive constraint (Context $\to$ Target), BiJEPA treats the relationship between data views $x$ and $y$ symmetrically \footnote{This symmetry, however, can be weighted, as discussed later in loss design.}. To capture the full semantic structure of the data, we introduce two distinct predictors: a forward predictor $P_{fwd}$ and a backward predictor $P_{bwd}$.

\subsubsection{Symmetric Prediction}
The BiJEPA architecture, as shown in Fig.\ref{fig:arch_comparison}, operates via two concurrent prediction cycles:

\paragraph{Forward pass ($x \to y$):}
this pass mimics the standard JEPA objective. The Online Encoder processes $x$ to produce $s_x$, and the \textit{Forward Predictor} attempts to predict the representation of $y$ ($s_y$) generated by the Target Encoder:
\begin{equation}
    \hat{s}_y = P_{fwd}(s_x), \quad \text{where} \quad s_x = f_\theta(x), \; s_y = f_{\bar{\theta}}(y).
\end{equation}

\paragraph{Backward pass ($y \to x$):}
simultaneously, the model learns the \textit{inverse mapping}. The Online Encoder processes $y$ to produce $s_y$, and the \textit{Backward Predictor} attempts to reconstruct the representation of $x$ ($s_x$) generated by the Target Encoder:
\begin{equation}
    \hat{s}_x = P_{bwd}(s_y), \quad \text{where} \quad s_y = f_\theta(y), \; s_x = f_{\bar{\theta}}(x).
\end{equation}
Note that in the backward pass, $y$ serves as the input to the gradient-receiving Online Encoder, while $x$ serves as the regression target via the EMA Target Encoder.

\subsubsection{Asymmetric Data-Weighted Loss}
While the architecture is structurally symmetric, the information content or availability of $x$ and $y$ may differ. To accommodate this, we define the total objective as a weighted convex combination of the forward and backward errors:
\begin{equation} \label{eq:BiJEPA_loss}
    L_{total} = \alpha ||\hat{s}_y - s_y||^2_2 + (1 - \alpha) ||\hat{s}_x - s_x||^2_2
\end{equation}
where $\alpha \in [0, 1]$ is a scalar weighting coefficient. This generalized formulation encapsulates the standard uni-directional JEPA as a special case where $\alpha = 1$, effectively disabling the backward predictor. In standard symmetric training, we set $\alpha = 0.5$. However, this flexibility allows for \textit{asymmetric data weighting}, where $\alpha$ can be derived dynamically based on the proportion of available information in $x$ versus $y$ (e.g., if view $y$ is significantly sparser or noisier, $\alpha$ can be adjusted to prioritize the more reliable forward signal).

\subsection{Stability Mechanism: Norm Regularization}
A critical finding during our methodology development was the phenomenon of \textit{Representation Explosion}. In a bi-directional setting, the feedback loops between the Online and Target encoders are amplified. Without constraints, the optimization landscape encourages the encoders to scale embedding vectors to infinity to minimize relative error. To mitigate this, we explore two regularization strategies:

\paragraph{Hard Constraint (Unit Sphere).} We can enforce a hard spherical constraint by projecting all embeddings onto a unit hypersphere:
\begin{equation}
    s = \frac{f(x)}{||f(x)||_2}
\end{equation}
This guarantees stability by bounding the vector norm to 1.0. However, as shown in our experiments (Appendix \ref{app:sphere_norm}), this removes the vector magnitude as a carrier of information (e.g., signal amplitude), potentially limiting representation capacity.

\paragraph{Soft Constraint (Expressive).} Alternatively, we can rely on soft constraints such as \textit{Layer Normalization} combined with \textit{Weight Decay}. This prevents unbounded growth while allowing the model to utilize vector magnitude to encode semantic intensity. We refer to this as the "Expressive" configuration and utilize it for our primary results.

\subsection{Inference and Model Usage}

After the self-supervised pre-training phase, the Target Encoder ($f_{\bar{\theta}}$) is discarded. The Online Encoder ($f_\theta$) acts as the primary feature extractor. 
At inference time, BiJEPA offers two distinct modes of operation:
\begin{enumerate}
    \item \textbf{Discriminative tasks:} for downstream applications such as classification, the input $x$ is mapped to its latent representation $s_x$. Note that if training utilized the hard constraint, $s_x$ must be normalized during inference; for the soft constraint configuration, the standard encoder output is used.
    
    \item \textbf{Generative imputation:} unlike standard uni-directional models, BiJEPA retains both trained predictors, $P_{fwd}$ and $P_{bwd}$. This enables bi-directional latent planning: the model can forecast future states from current observations via $P_{fwd}$, or infer unobserved past causes from current effects via $P_{bwd}$, effectively serving as a reversible world model.
\end{enumerate}

% ======================================================================
% MAIN TEXT
% ======================================================================

\section{Experiments}
\label{sec:exp}

We evaluated BiJEPA on datasets of increasing complexity. The primary goals of these experiments were to: (1) verify the stability mechanics of the symmetric training objective, (2) assess the quality of the learned representations for discriminative tasks, and (3) evaluate the model's capacity for generative forecasting in latent space.

\subsection{Evaluation Protocol}
To comprehensively assess the learned representations, we employ two distinct evaluation protocols. These strategies allow us to distinguish between the information content of the static embeddings and the model's dynamic forecasting capabilities.

\subsubsection{Protocol A: Representation Quality (Encoder Probe)}
This protocol follows the standard self-supervised learning evaluation benchmark. After training BiJEPA, we freeze the weights of the Online Encoder $f_\theta$ and train a task-specific probe (e.g., a linear layer or lightweight MLP) directly on the context embedding $s_x = f_\theta(x)$ to predict the target $y$.
\begin{equation}
    \hat{y} = \text{Decoder}(s_x)
\end{equation}
This test evaluates the question: \textit{``does the current context representation contain distinct and extractable information about the future?''} It measures the semantic richness of the encoder's output \footnote{For sufficiency of Predictive State Representations (PSRs) in planning and control, see e.g. \cite{huang2026vjepavariationaljointembedding}.}.

\subsubsection{Protocol B: Generative Forecasting (Predictor Probe)}
This protocol evaluates BiJEPA as a \textit{generative world model}. Unlike Protocol A, we utilize the trained Forward Predictor $P_{fwd}$ to explicitly forecast the future latent state. We freeze both the Online Encoder and the Predictor, generate the predicted latent vector $\hat{s}_y$, and train a decoder to map this prediction to the observation space \footnote{This latent-observation relation is also observed in e.g. hidden Markov models, Kalman filters and control systems.}:
\begin{equation}
    x \xrightarrow{f_\theta} s_x \xrightarrow{P_{fwd}} \hat{s}_y \xrightarrow{\text{Decoder}} \hat{y}
\end{equation}
This test evaluates the question: \textit{``can the model explicitly predict the future state in latent space?''} This protocol is useful for verifying the dynamic consistency of the learned world model. It demonstrates that the predictor has learned the underlying causal dynamics (e.g., frequency and phase in a sine wave) rather than the encoder simply memorizing statistical correlations.

\subsection{Experiment 1: Synthetic Periodic Time Series (Sine Waves)}

\subsubsection{Objective and Setup}
The primary objective of this experiment was to isolate the "Representation Explosion" phenomenon and identify heuristic constraints required for stability. We utilized a synthetic dataset of noisy sine waves, where the task requires capturing temporal correlations to predict future states.

\paragraph{Synthetic Data Generation.}
Each sequence $S$ of length $T=20$ was sampled from the function $S(t) = \sin(\omega t + \phi) + \epsilon_t$, where frequency $\omega \sim \mathcal{U}(0.8, 1.2)$, phase $\phi \sim \mathcal{U}(0, 2\pi)$, and noise $\epsilon_t \sim \mathcal{N}(0, 0.05)$.
The data was processed in batches of size $N=64$. Each sequence was split into a \textbf{Context} ($t_{0} \dots t_{9}$) and a \textbf{Target} ($t_{10} \dots t_{19}$). Detailed experimental setup can be found in Appendix \ref{app:sphere_norm}.

\subsubsection{Results: Optimization vs. Stability}
\begin{figure}[t]
    \centering
    % Subfigure (a): Explosion
    \begin{subfigure}[b]{0.48\textwidth}
        \includegraphics[width=\linewidth]{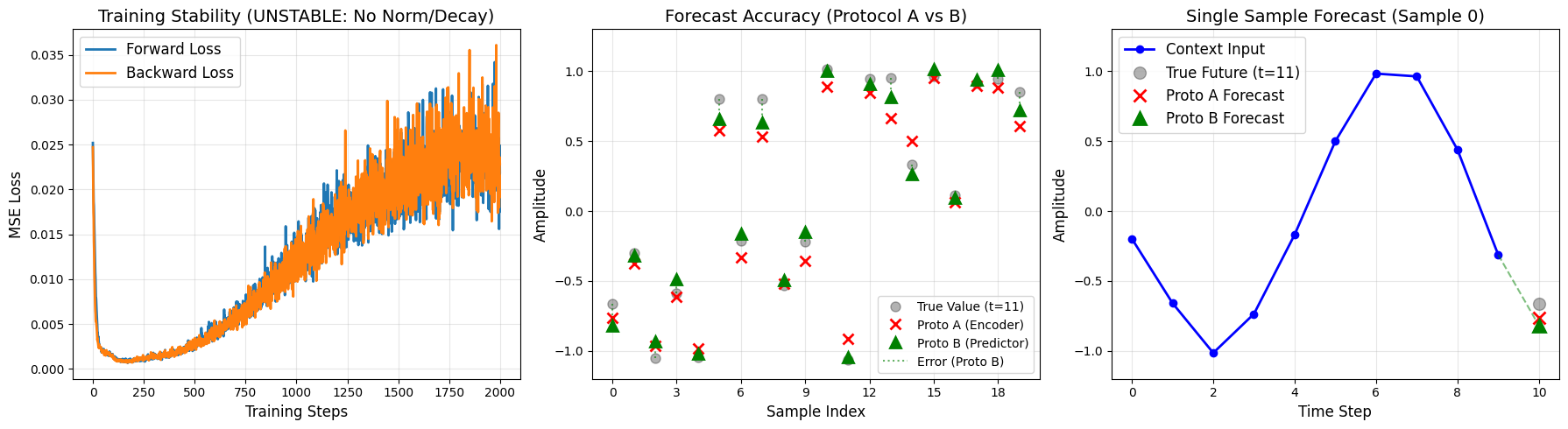}
        \caption{BiJEPA (Unconstrained)}
        \label{fig:sine_explosion}
    \end{subfigure}
    \hfill % Spacing
    % Subfigure (b): Expressive
    \begin{subfigure}[b]{0.48\textwidth}
        \includegraphics[width=\linewidth]{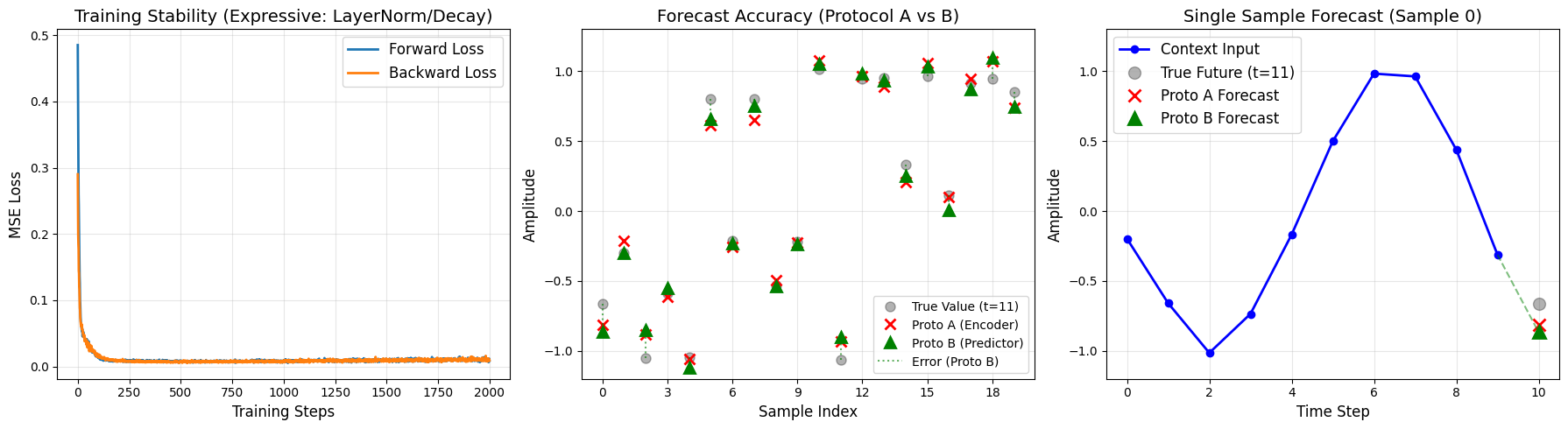}
        \caption{BiJEPA (Expressive)}
        \label{fig:sine_expressive}
    \end{subfigure}
    \hfill % Spacing
    % Subfigure (c): Classic JEPA
    \begin{subfigure}[b]{0.48\textwidth}
        \includegraphics[width=\linewidth]{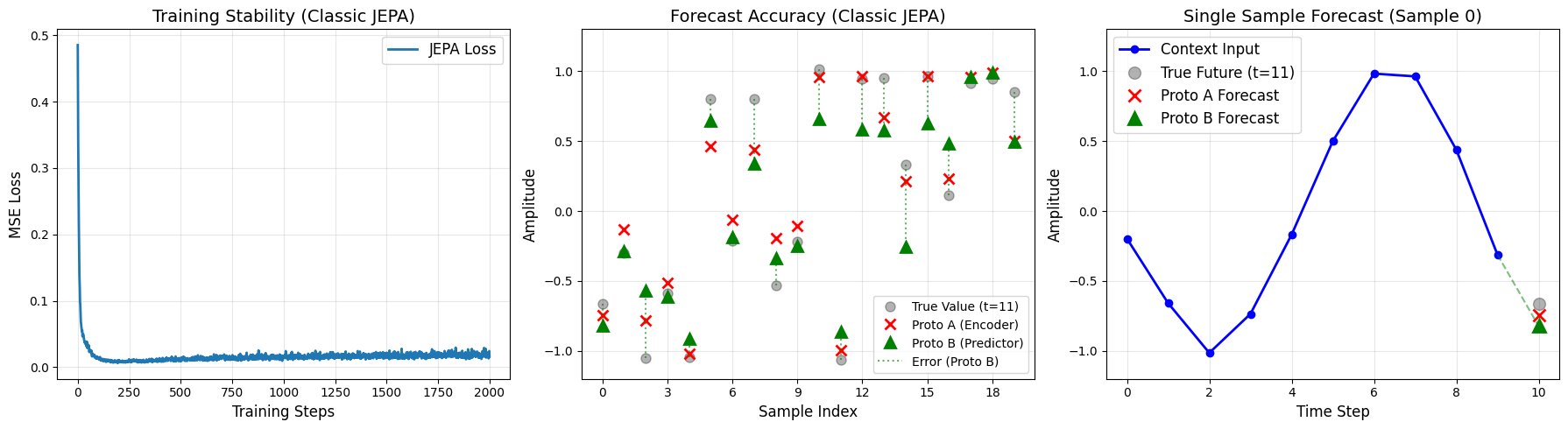}
        \caption{Classic JEPA (Benchmark)}
        \label{fig:sine_classic}
    \end{subfigure}
    \caption{\textbf{Impact of stability constraints \& architecture.} Each row displays: (Left) Training loss; (Middle) Batch forecasting accuracy (Sample Index vs Amplitude); (Right) Single-sample trajectory forecast. (a) Unconstrained BiJEPA diverges due to representation explosion. (b) Expressive BiJEPA achieves stable convergence and high accuracy. (c) Classic JEPA is stable but exhibits noisier loss dynamics and significantly higher forecasting error.}
    \label{fig:sine_comparison_main}
\end{figure}

We compared the training dynamics of an \textit{Unconstrained} baseline against our proposed \textit{Expressive} configuration to analyze the stability-plasticity trade-off.
In Fig. \ref{fig:sine_comparison_main}, each subfigure presents three views:
\begin{itemize}
    \item \textit{Left:} the training loss curves over 2,000 steps.
    \item \textit{Middle:} forecasting accuracy across a random batch of test samples, plotting the predicted amplitude (y-axis) against the sample index (x-axis) for both Protocol A (Encoder, red crosses) and Protocol B (Predictor, green triangles) and ground truth (grey circles).
    \item \textit{Right:} a single-sample trajectory showing the input context ($t_{0:9}$) and the 1-step forecast at $t_{10}$.
\end{itemize}

\paragraph{The Representation Explosion (Unconstrained).}
To empirically demonstrate the inherent instability of symmetric JEPA architectures, we trained a baseline model with all stability mechanisms removed (specifically, removing \textit{Layer Normalization} layers from the encoder and predictor architectures, and disabling \textit{Weight Decay} for all parameters during optimization).
As shown in Fig. \ref{fig:sine_comparison_main}(a), the training dynamics exhibit a characteristic failure mode. While the model initially learns the signal structure (loss drops to $0.002$ at step 500), the feedback loop between forward and backward predictors causes the latent vector magnitudes to grow unbounded. This leads to a divergence phase where the loss rebounds significantly ($0.002 \to 0.019$ by step 1500). 
Interestingly, the generative forecasting (Middle panel) remains accurate (Protocol B MSE: 0.0068) because the linear probe acts as an adaptor: it learns extremely small weights to rescale the massive (but structurally preserved) embeddings back to the observation range. However, despite this temporary success, the rising loss confirms that the optimization is mathematically divergent.

\paragraph{Stable Convergence with Soft Constraints (Expressive).}
We found that rigid constraints (e.g. projecting to the unit sphere, see Appendix \ref{app:sphere_norm}) are not strictly necessary for stability if appropriate "soft" constraints are applied. By incorporating \textit{Layer Normalization} and \textit{Weight Decay} ($\lambda=1e^{-4}$), the model achieves stable convergence (Fig. \ref{fig:sine_comparison_main}(b)).
The loss stabilizes at $\approx 0.009$, and the generative forecasting error remains low (Protocol B MSE: \textbf{0.013}). The middle and right panels confirm that the predictor (green) closely tracks the ground truth (grey) without the numerical instability seen in the unconstrained model.

\paragraph{Benchmark: Bi-Directional vs. Classic JEPA.}
Critically, our BiJEPA architecture significantly outperforms the standard uni-directional Classic JEPA. Even when using the same "Expressive" configuration, Classic JEPA achieves a much higher forecasting error (Protocol B MSE 0.052 vs 0.013, see Table \ref{tab:ablation_comparison}).
Furthermore, by comparing the loss curves (Left panels of (b) and (c)), we observe that \textbf{BiJEPA exhibits fewer fluctuations} than the Classic JEPA baseline. This suggests that the bi-directional consistency check ($x \leftrightarrow y$) creates a smoother optimization landscape, acting as a regularizer that prevents the transient instabilities often seen in uni-directional forcing.

\subsection{Experiment 2: Chaotic Dynamics (Lorenz Attractor)} \label{sec:Lorenz}

\subsubsection{Objective and Setup}
While sine waves represent simple periodic motion, real-world physical systems are often chaotic. Here, we investigate the \textbf{Lorenz Attractor}, a system of ordinary differential equations known for its sensitivity to initial conditions. The goal of this experiment was to test if BiJEPA can learn a stable latent model of non-linear, chaotic dynamics where small errors grow exponentially.

\paragraph{Data Generation.}
We generated 3D trajectories $(x, y, z)$ using the standard Lorenz equations:
\begin{equation}
    \begin{aligned}
        \frac{dx}{dt} &= \sigma(y - x) \\
        \frac{dy}{dt} &= x(\rho - z) - y \\
        \frac{dz}{dt} &= xy - \beta z
    \end{aligned}
    \label{eq:lorenz}
\end{equation}
We utilized standard chaotic parameters $\sigma=10, \rho=28, \beta=8/3$ with a time step of $dt=0.01$. To ensure diverse trajectory coverage within the attractor, initial conditions $(x_0, y_0, z_0)$ were sampled uniformly from $\mathcal{U}(-15, 15)$. The resulting data was normalized to zero mean and unit variance. Similar to Experiment 1, we used a context window of $T=20$ steps to predict the subsequent $T=20$ steps.

\subsubsection{Results: Modeling Chaos}
\begin{figure}[t]
    \centering
    % BiJEPA Result
    \begin{subfigure}[b]{0.48\textwidth}
        \includegraphics[width=\linewidth]{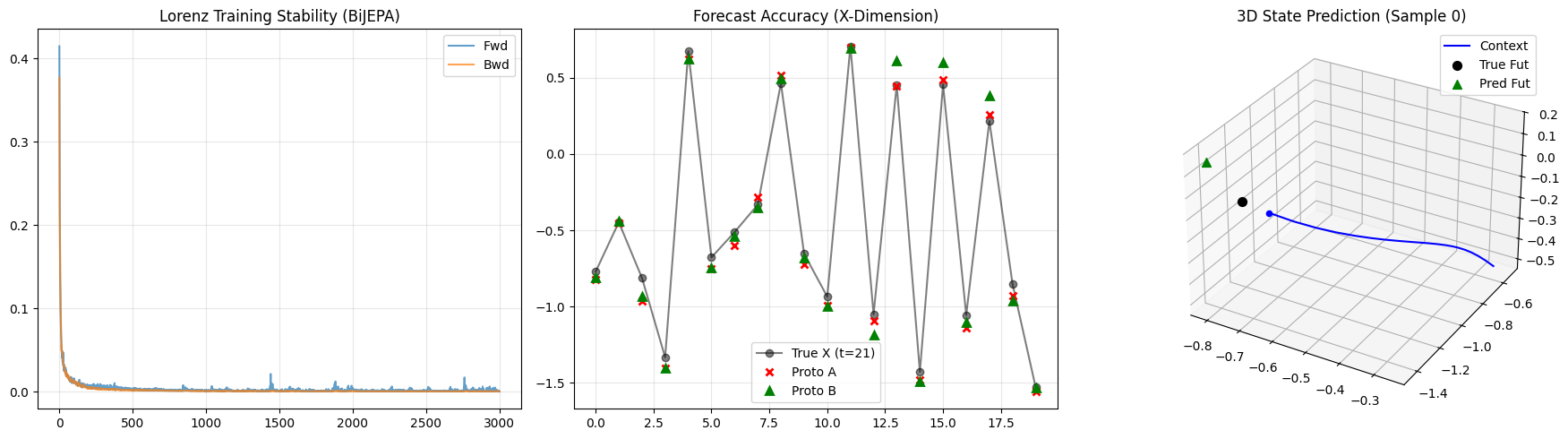}
        \caption{BiJEPA (Ours)}
        \label{fig:lorenz_bijepa}
    \end{subfigure}
    \hfill
    % Classic JEPA Result
    \begin{subfigure}[b]{0.48\textwidth}
        \includegraphics[width=\linewidth]{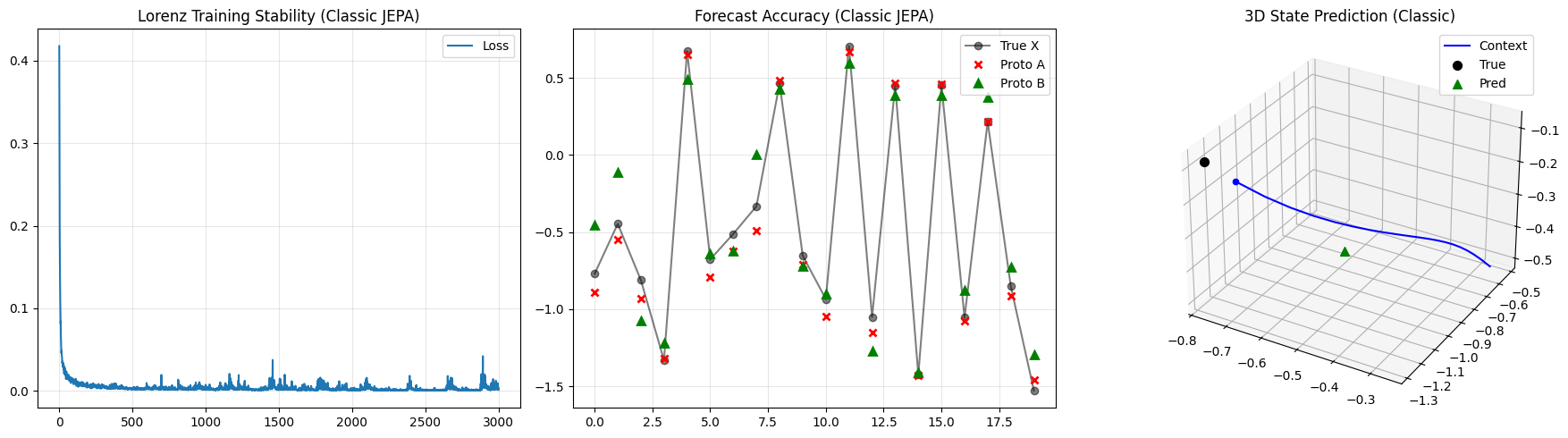}
        \caption{Classic JEPA (Baseline)}
        \label{fig:lorenz_classic}
    \end{subfigure}
    \caption{\textbf{Forecasting chaotic dynamics.} (Left) Training loss curves showing BiJEPA's superior stability. (Middle) 1-step forecast accuracy for the X-coordinate across a batch of 20 random test samples; BiJEPA (green) tracks the truth (grey) better than Classic JEPA (Red). (Right) 3D phase space reconstruction of a single sample trajectory.}
    \label{fig:lorenz_comparison}
\end{figure}

We compared the \textit{Expressive BiJEPA} against the \textit{Classic JEPA} baseline. Both models were trained for 3000 steps. Table \ref{tab:lorenz_results} summarizes the performance, and Fig.\ref{fig:lorenz_comparison} visualizes the training dynamics, batch forecasting accuracy, and phase-space reconstruction.

\begin{table}[ht]
    \centering
    \caption{\textbf{Lorenz Attractor Results.} BiJEPA significantly outperforms Classic JEPA in generative forecasting (Protocol B), reducing error by nearly \textbf{4x}.}
    \label{tab:lorenz_results}
    \resizebox{\columnwidth}{!}{%
    \begin{tabular}{lccc}
        \toprule
        \textbf{Model} & \textbf{Train Loss} & \textbf{Proto A} & \textbf{Proto B} \\
         & \textit{(Self-Sup)} & \textit{(Encoder)} & \textit{(Predictor)} \\
        \midrule
        \textbf{Classic JEPA} & 0.0006 & 0.0224 & 0.0937 \\
        \textbf{BiJEPA (Ours)} & 0.0007 & \textbf{0.0172} & \textbf{0.0249} \\
        \bottomrule
    \end{tabular}%
    }
\end{table}

\paragraph{Optimization stability.}
Comparing the training loss curves (Fig.\ref{fig:lorenz_comparison}, Left), BiJEPA demonstrates superior stability. While the Classic JEPA loss fluctuates visibly throughout training, the BiJEPA loss decreases monotonically with very small fluctuations. This suggests that the symmetric objective creates a smoother optimization landscape, preventing the transient instabilities often seen in uni-directional forcing.

\paragraph{The failure of uni-directional prediction.}
The Classic JEPA baseline exhibits a common failure mode in chaotic modeling. Although it achieves a low training loss, its generative forecasting capability is poor (Protocol B MSE: 0.0937).
We visualized the 1-step forecast accuracy across 20 diverse test samples (Fig.\ref{fig:lorenz_comparison}, Middle). Classic JEPA (red crosses) frequently misses the exact coordinate values (grey circles). Furthermore, the 3D reconstruction of a single sample trajectory (Fig.\ref{fig:lorenz_comparison}, Right) shows that the model captures the general attractor shape but fails to pinpoint the exact future state on the manifold. By minimizing a uni-directional loss, the model likely collapses towards a mean-field prediction that smooths over the precise chaotic details.

\paragraph{Bi-directional consistency enforces precision.}
In contrast, BiJEPA achieves a forecasting error of \textbf{0.0249}, roughly \textbf{3.7x lower} than the baseline.
The middle plot confirms that BiJEPA (green triangles) tracks the ground truth across the entire batch of samples with high precision. Similarly, the single-sample trajectory reconstruction (Right) shows tight alignment between the predicted and actual future states.
We hypothesize that the symmetric consistency check ($X \to Y$ and $Y \to X$) acts as a strong regularizer. It forces the latent space to respect the \textit{reversibility} of the underlying ODEs (to the extent possible), preventing the model from taking "shortcuts" that satisfy the forward loss but violate the backward dynamics.

\subsection{Experiment 3: Spatial Vision (MNIST)}

\subsubsection{Objective and Setup}
To verify that BiJEPA's benefits extend beyond temporal dynamics to spatial coherence, we evaluated it on the MNIST dataset.
\textbf{Setup:} we split each $28 \times 28$ digit image vertically. The \textit{Context} was the left half ($14 \times 28$), and the \textit{Target} was the right half. The model must learn to infer the missing structure (e.g., the loop of a '6' or the stem of a '7') solely from the partial view.

\subsubsection{Results: Spatial Inpainting \& Classification}
We compared the representation quality of the \textit{Classic JEPA} (uni-directional) against the \textit{BiJEPA} (bi-directional) using two protocols:
\begin{enumerate}
    \item \textbf{Discriminative probe:} a linear classifier trained on the frozen "Left Half" embeddings to predict the digit class (0-9).
    \item \textbf{Generative decoder:} a decoder trained to map the "Left Half" embedding to the pixels of the "Right Half" (Hallucination).
\end{enumerate}

Table \ref{tab:mnist_results} summarizes the results.

\begin{table}[ht]
    \centering
    \caption{\textbf{Spatial Vision Results (MNIST).} BiJEPA outperforms the Classic JEPA baseline in both classification accuracy and generative reconstruction quality, reducing the MSE and boosting accuracy by over 2.7\%.}
    \label{tab:mnist_results}
    \resizebox{\columnwidth}{!}{%
    \begin{tabular}{lcc}
        \toprule
        \textbf{Model} & \textbf{Linear Probe Acc.} & \textbf{Decoder MSE} \\
         & \textit{(Discriminative)} & \textit{(Generative)} \\
        \midrule
        \textbf{Classic JEPA} & 89.14\% & 0.0305 \\
        \textbf{BiJEPA (Ours)} & \textbf{91.88\%} & \textbf{0.0298} \\
        \bottomrule
    \end{tabular}%
    }
\end{table}

\paragraph{Superior semantic abstraction.}
The linear probe accuracy is a proxy for how linearly separable the classes are in the latent space. BiJEPA achieves 91.88\% accuracy, significantly outperforming the Classic JEPA (89.14\%).
This result is non-trivial; classifying digits from only their left half is ambiguous (e.g., a '3' and an '8' often look similar on the left). The higher accuracy suggests that BiJEPA's backward constraint ($Right \to Left$) forces the encoder to capture more subtle, global structural cues that a one-way predictor might ignore.

\begin{figure*}[t]
    \centering
    \begin{subfigure}[t]{0.45\textwidth}
        \centering
        \includegraphics[
            height=0.35\textheight,
            keepaspectratio
        ]{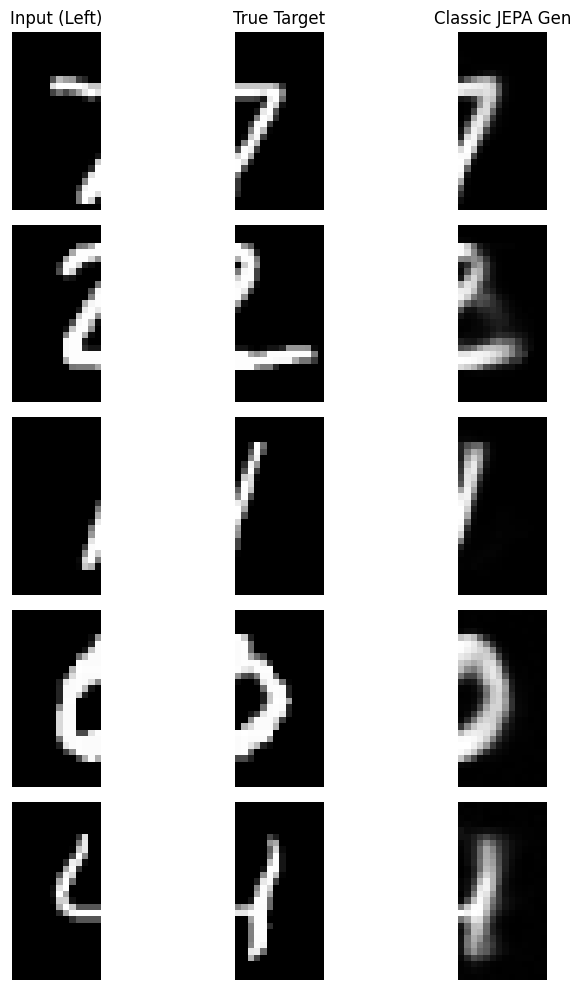}
        \caption{Classic JEPA (89.14\% Acc)}
        \label{fig:mnist_classic}
    \end{subfigure}
    \hfill
    \begin{subfigure}[t]{0.45\textwidth}
        \centering
        \includegraphics[
            height=0.35\textheight,
            keepaspectratio
        ]{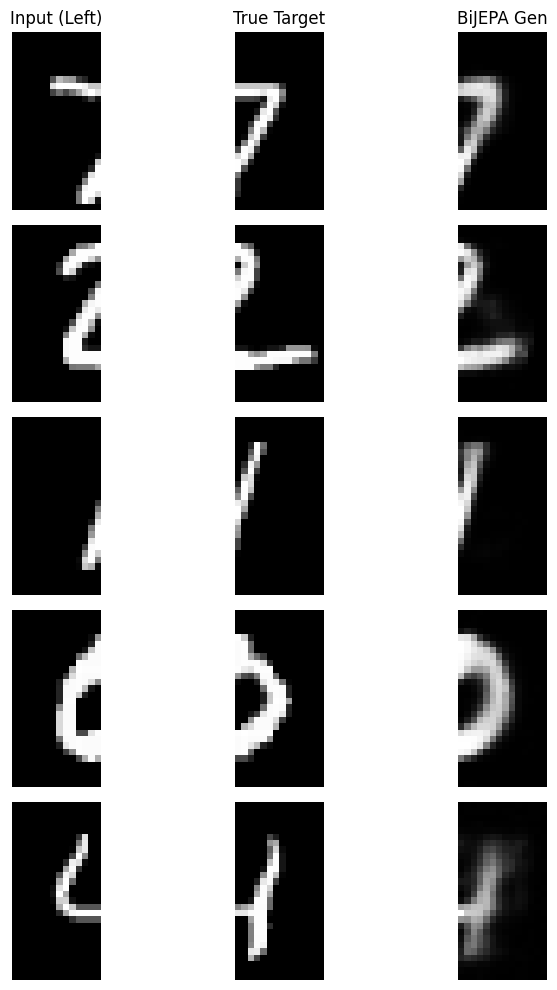}
        \caption{BiJEPA (91.88\% Acc)}
        \label{fig:mnist_bijepa}
    \end{subfigure}

    \caption{\textbf{Generative hallucinations.} The models are given only the \textit{Left Half} (Input) and must generate the \textit{Right Half}. BiJEPA (b) produces sharper and more semantically consistent completions than the baseline (a), particularly for difficult digits like '2' and '4'.}
    \label{fig:mnist_comparison}
\end{figure*}

\paragraph{Generative "hallucination".}
To qualitatively assess what the model "knows," we visualized the output of the generative decoder. As shown in Fig.\ref{fig:mnist_comparison}, the model successfully "hallucinates" the missing right half of the digits.
For example, given the left half of a '2' (Row 2), the model correctly infers the bottom curve and the top arch. Importantly, the reconstructions are not just blurry averages; they possess correct geometric structure, confirming that the latent state $s_x$ encodes high-level shape semantics rather than just texture statistics.

% ----------------------------------------------------------------------
\section{Discussion}
\label{sec:discuss}

\subsection{The Power of Bi-Directionality}
The primary advantage of BiJEPA lies in its rigorous \textit{enforcement of semantic consistency}. Standard uni-directional models ($x \to y$) often learn "shortcut" features, i.e. statistical correlations that help predict $y$ without truly understanding the structure of $x$. By enforcing the inverse mapping ($y \to x$), BiJEPA acts as a powerful \textit{regularizer}. It demands that the representation $s_y$ retains enough information to reconstruct the semantic content of $s_x$, preventing information collapse. This was empirically validated in the MNIST experiment, where the backward constraint ($Right \to Left$) significantly improved the classification accuracy of the forward embeddings, suggesting a more \textit{globally} coherent latent structure.

\subsection{Architectural Variations: The Shared Predictor} 
\label{subsec:shared_predictor}
In our current implementation, $P_{fwd}$ and $P_{bwd}$ are distinct neural networks. However, for systems governed by strictly reversible conservation laws (e.g., Hamiltonian mechanics), this separation may be redundant. A potential optimization is the use of \textit{Invertible Neural Networks (INNs)} \cite{ehrlich2026pseudoinvertibleNN} or Normalizing Flows. By designing a single, bijective predictor $P$ such that $s_y = P(s_x)$ and $s_x = P^{-1}(s_y)$, we could halve the parameter count of the predictive head and enforce mathematical invertibility by construction. This remains a promising avenue for tasks requiring strict physical plausibility.

\subsection{The Mechanics of Representation Explosion}
A central theoretical finding of this work is that \textit{symmetric} predictive architectures are inherently unstable. As observed in Experiment 1 (Fig.\ref{fig:sine_comparison_main}), unconstrained BiJEPA models suffer from \textit{Representation Explosion}. This phenomenon arises from the compounding interaction of two dynamics:
\begin{enumerate}
    \item \textbf{The EMA chasing game:} the Target Encoder ($f_{\bar{\theta}}$) tracks the Online Encoder ($f_\theta$). If the Online Encoder's output magnitude increases, the Target follows. To minimize distance in the next step, the Online Encoder must match this larger target, creating a continuous expansion cycle.
    \item \textbf{Bi-directional feedback:} unlike uni-directional models where the target is passive, BiJEPA couples two active optimization loops. A perturbation in $s_x$ amplifies $s_y$, which in turn amplifies $s_x$ via the backward loss.
\end{enumerate}
Mathematically, the predictor-encoder loop acts as a linear system $h_{t+1} \approx W h_t$. Without constraints, the optimization landscape allows the eigenvalues of $W$ to exceed 1.0, leading to unbounded signal growth. Our results confirm that \textbf{norm regularization} is not merely a regularization trick but a structural necessity. While hard-clamping (L2-sphere) guarantees stability, our experiments favor soft constraints (LN + Decay) to balance stability with representation capacity.

\subsection{Semantic vs. Metric Representations}
The Lorenz experiment (Sec.\ref{sec:Lorenz}) highlighted a key distinction between JEPA and reconstruction-based methods (e.g., MAE). As shown in Table.\ref{tab:lorenz_results}, BiJEPA achieved near-perfect \textit{latent} predictability (Train Loss $\approx 0$) while maintaining a non-zero metric error in the probe (MSE $\approx 0.02$). This indicates that BiJEPA is a \textit{semantic extractor}, not a compressor. It prioritizes information that is temporally predictable (the attractor's phase and topology) while discarding high-frequency metric noise. This property makes it ideal for hierarchical planning, where abstract state dynamics are more valuable than pixel-perfect reconstruction.

\subsection{Asymmetric and Adaptive Loss Weighting}
While we utilized a balanced loss ($\alpha=0.5$ in Eq.\ref{eq:BiJEPA_loss}), real-world data is often asymmetric. In causal sequence modeling, the past ($x$) is fixed, while the future ($y$) is uncertain. In cases of occlusion, one view may contain significantly less information than the other.
BiJEPA supports \textbf{asymmetric data-weighted loss}, where $\alpha$ can be dynamic:
\begin{equation}
    \alpha = \frac{I(x)}{I(x) + I(y)}
\end{equation}
where $I(\cdot)$ is an estimate of information content (e.g. size of data) or sparsity. Future implementations could dynamically adjust $\alpha$ to down-weight the loss contribution of occluded or noisy views, preventing the model from hallucinating features in the backward pass that simply do not exist in the corrupted target.

\subsection{Future Work}

\paragraph{Stochasticity and uncertainty.}
The current BiJEPA formulation is deterministic: $P(s_x) \to \hat{s}_y$. However, many systems (like the Lorenz attractor near bifurcation points) are inherently stochastic or multi-modal. Future work should explicitly utilize the latent conditioning variable $z$ (currently unused in simple implementations) to inject noise, transforming the predictor into a generative model $P(s_x, z) \to \hat{s}_y$ capable of outputting a distribution of possible futures rather than a single mean-field prediction.

\paragraph{Scaling and applications.}
We aim to scale BiJEPA to high-resolution video using Vision Transformers (ViT). The learned bi-directional dynamics have immediate applications in \textit{model-based Reinforcement Learning}, where an agent could use $P_{fwd}$ to plan actions and $P_{bwd}$ to infer unobserved causes of failure (counterfactual reasoning). 
Beyond robotics, the symmetric architecture is uniquely suited for \textit{inverse molecular design}, where the forward pass predicts protein structure from sequence ($x \to y$) and the backward pass generates novel sequences for target structures ($y \to x$). Similarly, in \textit{high-fidelity video interpolation}, BiJEPA can enforce cycle-consistency between frames $t$ and $t+k$ to hallucinate coherent intermediate states. 
In \textit{recommender systems}, BiJEPA could model the symmetric interaction between users and items: predicting item affinity from user history ($User \to Item$) while simultaneously profiling target user demographics from item characteristics ($Item \to User$).
Further, the stability provided by L2-normalization makes BiJEPA a strong candidate for \textit{Sim-to-Real} transfer\footnote{In Sim-to-Real transfer, agents face a ``reality gap'' where real-world inputs (e.g., camera feed) differ in lighting or texture intensity from the simulator. Unconstrained models often encode these variations as shifts in vector magnitude, causing the representation to drift out of the training distribution. L2-normalization enforces scale invariance, ensuring that the model relies solely on vector direction (semantic structure) rather than signal intensity, thereby preventing catastrophic failure when environmental conditions change.} and \textit{anomaly detection}, where physical violations can be flagged by high forward-backward inconsistency error.

\section{Conclusion}
\label{sec:conc}

In this paper, we introduced \textbf{BiJEPA}, a symmetric framework for self-supervised representation learning. By extending the Joint Embedding Predictive Architecture to explicitly model bi-directional consistency, we address the limitations of standard uni-directional pre-training. 

Our key contributions are three-fold:
\begin{enumerate}
    \item \textbf{Architecture:} we demonstrated that adding a backward predictor captures richer semantic information and enhances predictive consistency, improving downstream classification accuracy on MNIST by 2.7\% over the Classic JEPA baseline.
    \item \textbf{Stability:} we identified \textit{Representation Explosion} as the fundamental failure mode of symmetric self-supervision and demonstrated that effective norm constraints (specifically Layer Normalization and Weight Decay) are necessary to ensure convergence without sacrificing representation capacity.
    \item \textbf{Dynamics:} through the Lorenz attractor experiment, we showed that BiJEPA learns a more precise internal model of chaotic dynamics (nearly 4x lower forecasting error) compared to forward-only models, which tend to collapse to mean-field approximations.
\end{enumerate}

BiJEPA offers a more holistic approach to world modeling - one that respects the physical reversibility of time and space. We hope this work serves as a foundation for future research into physically consistent, bi-directional foundation models for robotics, generative media, and scientific discovery.

% ----------------------------------------------------------------------
\bibliographystyle{plain}
\bibliography{reference}

\appendix
\section{Ablation: The Cost of Hard Constraints}
\label{app:sphere_norm}

In the main text, we utilized an "Expressive" configuration (soft constraint on embedding vector via heuristics such as \textit{LayerNorm} + \textit{Decay}) that allowed embedding magnitudes to vary. Here, we quantitatively compare this against the "Restrictive" (Unit Sphere) configuration, the "Unstable" (Explosion) baseline, and a standard "Classic JEPA" benchmark.

\begin{figure}[ht]
    \centering
    \includegraphics[width=0.9\linewidth]{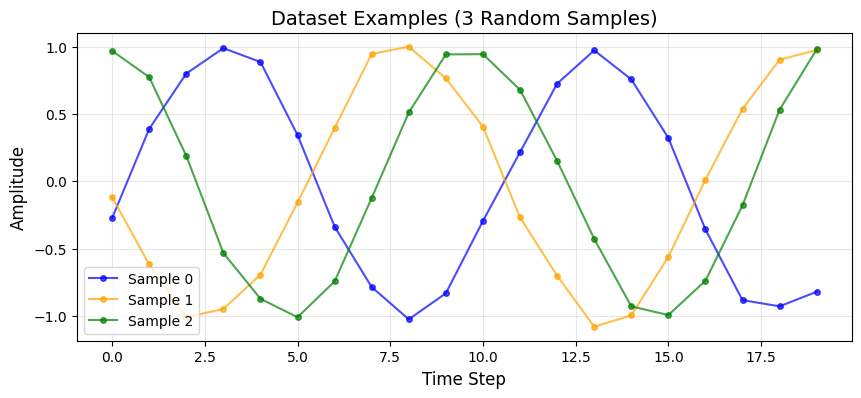}
    \caption{\textbf{Dataset samples.} Three random sine waves from the validation set showing variations in frequency $\omega \sim \mathcal{U}(0.8, 1.2)$ and phase $\phi \sim \mathcal{U}(0, 2\pi)$.}
    \label{fig:sine_samples}
\end{figure}

\paragraph{Data Complexity.} 
To understand the difficulty of the forecasting task, Fig.\ref{fig:sine_samples} presents three random samples from the dataset. The model must infer the distinct frequency and phase from the context window ($t_{0:9}$) to accurately predict the target trajectory ($t_{10:19}$).

\paragraph{Quantitative Comparison.}
Table \ref{tab:ablation_comparison} summarizes the performance metrics across all four configurations. 

\begin{table}[ht]
    \centering
    \caption{\textbf{Quantitative Comparison of Stability Mechanisms \& Baselines.} We report final training loss and MSE for both protocols. The \textit{BiJEPA Expressive} model offers the optimal trade-off, outperforming both the restrictive baseline and the classic uni-directional architecture.}
    \label{tab:ablation_comparison}
    % Resize the table to fit exactly within the column width
    \resizebox{\columnwidth}{!}{%
    \begin{tabular}{lcccc}
        \toprule
        \textbf{Configuration} & \textbf{Stability} & \textbf{Train Loss} & \textbf{Proto A} & \textbf{Proto B} \\
         & & & \textit{(Encoder)} & \textit{(Predictor)} \\
        \midrule
        \textbf{BiJEPA Explosion} (None) & Divergent & 0.0190 (Rise) & 0.0148 & \textbf{0.0068} \\
        \textbf{BiJEPA Expressive} (LN+Decay) & \textbf{Stable} & 0.0090 (Flat) & \textbf{0.0093} & 0.0132 \\
        \textbf{BiJEPA Restrictive} (Sphere) & \textbf{Stable} & \textbf{0.0004} (Flat) & 0.1460 & 0.0779 \\
        \midrule
        \textbf{Classic JEPA} (LN+Decay) & Stable & 0.0143 (Flat) & 0.0279 & 0.0517 \\
        \bottomrule
    \end{tabular}%
    }
\end{table}

% \textbf{Analysis:}
\begin{itemize}
    \item \textbf{Explosion vs. Expressive:} the Unconstrained model achieves the lowest Predictor error (0.0068), confirming the optimization landscape is deep, but its training loss diverges. The Expressive model sacrifices minimal precision (0.0132) for guaranteed stability.
    \item \textbf{Expressive vs. Restrictive:} enforcing the unit sphere constraint (Restrictive) reduces the training loss near zero ($4e^{-4}$) because the target space is bounded. However, this paradoxically hurts downstream performance (Predictor MSE rises to 0.0779), as the model loses the ability to encode signal amplitude in the vector magnitude.
    \item \textbf{BiJEPA vs. Classic JEPA:} the Classic JEPA is stable but significantly underperforms the BiJEPA Expressive model (Protocol B MSE 0.052 vs 0.013). This highlights that the symmetric, bi-directional objective forces the encoder to learn a more robust and predictive representation of the underlying dynamics.
\end{itemize}

\begin{figure}[ht]
    \centering
    \includegraphics[width=1.0\linewidth]{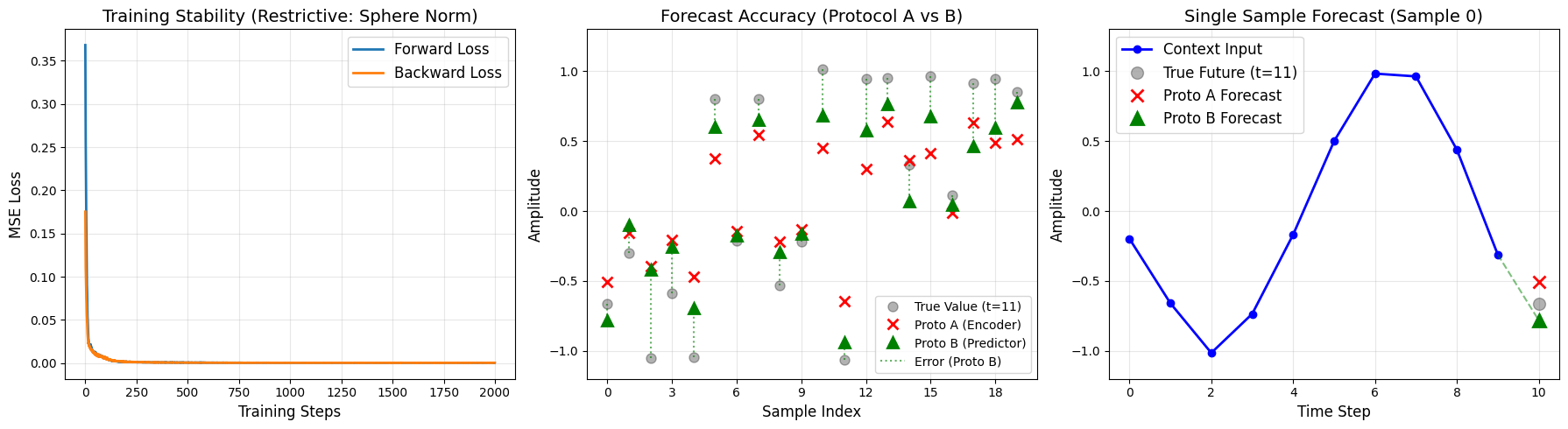}
    \caption{\textbf{Restrictive configuration (unit sphere) results.} While training is perfectly stable (Left), the forecasting accuracy (Middle) is visibly lower than the Expressive model, with larger deviations from the ground truth. The train loss curves, similar to Expressive BiJEPA, are smoother than that of Classic JEPA in Fig.\ref{fig:sine_comparison_main}.}
    \label{fig:sine_restrictive}
\end{figure}

\section{Experimental Environment and Setup}
\label{app:setup}

\subsection{Hardware and Software Environment}
All experiments were conducted on a Google Colab instance \cite{Edwards2024Colab} with following specifications:
\begin{itemize}
    \item \textbf{CPU:} Intel(R) Xeon(R) CPU @ 2.20GHz (1 physical core, 2 threads per core).
    \item \textbf{Memory:} 13.61 GB System RAM.
    \item \textbf{Disk Storage:} 242.49 GB.
    \item \textbf{Acceleration:} experiments were accelerated using NVIDIA T4 GPUs where available (standard Colab GPU runtime).
    \item \textbf{Software:} Python 3.10, PyTorch 2.1+, CUDA 12.1.
\end{itemize}

\subsection{Data Generation and Processing}

\paragraph{Experiment 1: Synthetic Periodic Signals (Sine Waves).}
Data was generated on-the-fly during training to prevent overfitting to a fixed set of samples.
\begin{itemize}
    \item \textbf{Function:} $S(t) = \sin(\omega t + \phi) + \epsilon_t$.
    \item \textbf{Parameters:} Frequency $\omega \sim \mathcal{U}(0.8, 1.2)$, Phase $\phi \sim \mathcal{U}(0, 2\pi)$.
    \item \textbf{Noise:} Gaussian noise $\epsilon_t \sim \mathcal{N}(0, 0.05)$ added at each step.
    \item \textbf{Dimensions:} Sequence length $T=20$. Context $t_{0:9}$, Target $t_{10:19}$.
    \item \textbf{Batching:} Batch size $N=64$, generated fresh at every step.
\end{itemize}

\paragraph{Experiment 2: Chaotic Dynamics (Lorenz Attractor).}
Data was pre-generated and normalized to ensure consistent manifold structure across epochs.
\begin{itemize}
    \item \textbf{System Equations:} standard 3D Lorenz system ($\frac{dx}{dt} = \sigma(y-x)$, $\frac{dy}{dt} = x(\rho-z)-y$, $\frac{dz}{dt} = xy-\beta z$).
    \item \textbf{Parameters:} $\sigma=10, \rho=28, \beta=8/3$. Integration time step $dt=0.01$.
    \item \textbf{Initial Conditions:} starting states $(x_0, y_0, z_0)$ were sampled uniformly from $\mathcal{U}(-15, 15)$.
    \item \textbf{Preprocessing:} trajectories were normalized to zero mean and unit variance.
    \item \textbf{Dimensions:} sequence length $T=40$. Context $t_{0:19}$, Target $t_{20:39}$.
    \item \textbf{Dataset Size:} 2,000 training trajectories, 1,000 probe trajectories, 20 test trajectories.
\end{itemize}

\paragraph{Experiment 3: Spatial Vision (MNIST).}
We utilized the standard MNIST handwritten digit dataset.
\begin{itemize}
    \item \textbf{Preprocessing:} images ($28 \times 28$) were normalized with mean $0.1307$ and std $0.3081$.
    \item \textbf{Splitting:} vertical crop at column index 14.
    \item \textbf{Context:} left half ($28 \times 14$ pixels).
    \item \textbf{Target:} right half ($28 \times 14$ pixels).
    \item \textbf{Batching:} batch size $N=256$ with standard shuffling for training and fixed indices for visualization.
\end{itemize}

\subsection{Model Architectures and Hyperparameters}

\paragraph{Experiment 1 (Sine Waves).}
We utilized a lightweight MLP architecture for this low-dimensional task.
\begin{itemize}
    \item \textbf{Encoder:} Input (10) $\to$ Linear(64) $\to$ LN $\to$ ReLU $\to$ Linear(64) $\to$ LN $\to$ ReLU $\to$ Linear(16).
    \item \textbf{Predictor:} Input (16) $\to$ Linear(64) $\to$ LN $\to$ ReLU $\to$ Linear(16).
    \item \textbf{Optimizer:} AdamW, Learning Rate $10^{-3}$, Weight Decay $10^{-4}$.
    \item \textbf{Training:} 2,000 steps, Batch Size 64.
    \item \textbf{EMA Momentum ($\tau$):} 0.995.
\end{itemize}

\paragraph{Experiment 2 (Lorenz Attractor).}
To capture the complexity of chaotic dynamics, the hidden dimensions were increased.
\begin{itemize}
    \item \textbf{Encoder:} Input (60) $\to$ Linear(128) $\to$ LN $\to$ ReLU $\to$ Linear(128) $\to$ LN $\to$ ReLU $\to$ Linear(32).
    \item \textbf{Predictor:} Input (32) $\to$ Linear(128) $\to$ LN $\to$ ReLU $\to$ Linear(32).
    \item \textbf{Optimizer:} AdamW, Learning Rate $5 \times 10^{-4}$, Weight Decay $10^{-4}$.
    \item \textbf{Training:} 3,000 steps, Batch Size 64.
    \item \textbf{EMA Momentum ($\tau$):} 0.995.
\end{itemize}

\paragraph{Experiment 3 (MNIST).}
We employed a convolutional encoder to capture spatial structure.
\begin{itemize}
    \item \textbf{Encoder:}
    \begin{enumerate}
        \item Conv2d(1$\to$32, k=3, s=2, p=1), BN, ReLU.
        \item Conv2d(32$\to$64, k=3, s=2, p=1), BN, ReLU.
        \item Flatten $\to$ Linear(1792$\to$128), LN, ReLU $\to$ Linear(64).
    \end{enumerate}
    \item \textbf{Predictor:} Input (64) $\to$ Linear(128) $\to$ LN $\to$ ReLU $\to$ Linear(64).
    \item \textbf{Optimizer:} AdamW, Learning Rate $10^{-3}$, Weight Decay $10^{-4}$.
    \item \textbf{Training:} 10 Epochs, Batch Size 256.
    \item \textbf{EMA Momentum ($\tau$):} 0.99.
\end{itemize}

\subsection{Code Availability}
The complete source code and data generation procedures used for the experiments in this paper are available at: \url{https://github.com/YongchaoHuang/BiJEPA}.

\end{document}